\documentclass{article}

     \usepackage[preprint]{neurips_2024}

\usepackage[utf8]{inputenc} 
\usepackage[T1]{fontenc}    
\usepackage{hyperref}       
\usepackage{url}            
\usepackage{booktabs}       
\usepackage{amsfonts}       
\usepackage{nicefrac}       
\usepackage{microtype}      
\usepackage{xcolor}         

\usepackage[ruled,linesnumbered]{algorithm2e}
\usepackage{dsfont}
\usepackage{makecell}

\usepackage[utf8]{inputenc} 
\usepackage[T1]{fontenc}    
\usepackage{hyperref}       
\usepackage{url}            
\usepackage{booktabs}       
\usepackage{amsfonts}       
\usepackage{nicefrac}       
\usepackage{microtype}      
\usepackage{xcolor}         
\usepackage{amsmath}
\usepackage{amssymb}
\usepackage{mathtools}
\usepackage{amsthm}
\usepackage{bbm}
\usepackage[capitalize,noabbrev]{cleveref}
\usepackage{graphicx}
\usepackage{aliascnt}
\usepackage{float}
\usepackage{xr-hyper}
\usepackage{hyperref}
\usepackage{todonotes}
\usepackage{varwidth}
\usepackage{multicol}
\usepackage{changepage}
\usepackage{enumitem}
\usepackage{autonum}
\usepackage{xargs}
\usepackage[ruled,linesnumbered]{algorithm2e}

\makeatletter

\crefname{theorem}{theorem}{Theorems}
\Crefname{Theorem}{Theorem}{Theorems}

\newaliascnt{lemma}{theorem}

\aliascntresetthe{lemma}
\crefname{lemma}{lemma}{lemmas}
\Crefname{Lemma}{Lemma}{Lemmas}

\newaliascnt{corollary}{theorem}

\aliascntresetthe{corollary}
\crefname{corollary}{corollary}{corollaries}
\Crefname{Corollary}{Corollary}{Corollaries}

\newaliascnt{proposition}{theorem}

\aliascntresetthe{proposition}
\crefname{proposition}{proposition}{propositions}
\Crefname{Proposition}{Proposition}{Propositions}

\newaliascnt{definition}{theorem}

\aliascntresetthe{definition}
\crefname{definition}{definition}{definitions}
\Crefname{Definition}{Definition}{Definitions}

\Crefname{assumption}{\textbf{A}\hspace{-3pt}}{\textbf{A}\hspace{-3pt}}
\crefname{assumption}{\textbf{A}}{\textbf{A}}

\newaliascnt{remark}{theorem}

\aliascntresetthe{remark}
\crefname{remark}{remark}{remarks}
\Crefname{Remark}{Remark}{Remarks}

\crefname{example}{example}{examples}
\Crefname{Example}{Example}{Examples}

\crefname{algorithm}{algorithm}{algorithms}
\Crefname{Algorithm}{Algorithm}{Algorithms}

\crefname{figure}{figure}{figures}
\Crefname{Figure}{Figure}{Figures}

\usepackage{caption}
\usepackage{subcaption}
\usepackage{multirow}

\def\rset{\mathbb{R}}

\def\wrt{w.r.t.}

\def\Algo{\texttt{ReALLM}}

\def\ie{i.e.}
\def\eg{e.g.}
\def\nsteps{T}
\newcommandx{\M}[3][1=i,2=k,3=\nsteps]{\operatorname{M}^{#3}_{#1,#2}}
\newcommandx{\m}[3][1=i,2=k,3=\nsteps]{\operatorname{m}^{#3}_{#1,#2}}
\newcommandx{\mi}[1][1=i]{\operatorname{m}_{#1}}
\newcommandx{\weightm}[2][1=k,2=\nsteps]{\operatorname{m}^{#2}_{#1}}

\usepackage{xargs}
\newcommandx{\CPE}[3][1=]{\PE_{#1}\left[\left. #2 \, \right| #3 \right]}

\def\PE{\mathbb{E}}

\newcommandx{\indi}[2][1=]{\1^{#1}_{#2}}

\newcommand{\1}{\ensuremath{\mathds{1}}}

\title{\Algo: A general framework for LLM compression and fine-tuning}

\author{%
  Louis Leconte\thanks{equal contribution} \\
  Lisite, Isep, Sorbonne University\\
Math. and Algo. Sciences Lab, Huawei Tech\\
  \texttt{louis.leconte@ens-paris-saclay.fr} \\
  \And
  Lisa Bedin$^*$ \\
  CMAP, Ecole Polytechnique, France \\
  \texttt{lisa.bedin@polytechnique.edu} \\
  \AND
  Van Minh Nguyen \\
Math. and Algo.
Sciences Lab, Huawei Tech. \\
  \And
  Eric Moulines \\
  CMAP, Ecole Polytechnique, France \\
}

\begin{document}

\maketitle

\begin{abstract}
We introduce \Algo, a novel approach for compression and memory-efficient adaptation of pre-trained language models that encompasses most of the post-training quantization and fine-tuning methods for a budget of $<4$ bits. Pre-trained matrices are decomposed into a high-precision low-rank component and a vector-quantized latent representation (using an autoencoder). During the fine-tuning step, only the low-rank components are updated. Our results show that pre-trained matrices exhibit different patterns. \Algo\ adapts the shape of the encoder (small/large embedding, high/low bit VQ, etc.) to each matrix. \Algo\ proposes to represent each matrix with a small embedding on $b$ bits and a neural decoder model $\mathcal{D}_\phi$ with its weights on $b_\phi$ bits. The decompression of a matrix requires only one embedding and a single forward pass with the decoder. Our weight-only quantization algorithm yields the best results on language generation tasks (C4 and WikiText-2) for a budget of $3$ bits \emph{without} any training. With a budget of $2$ bits, \Algo\ achieves state-of-the art performance after fine-tuning on a small calibration dataset.
\end{abstract}

\section{Introduction}
Large Language Models (LLMs) based on transformer architectures \citep{vaswani2017attention} have attracted increasing interest, especially with the availability of high-quality, open-source LLMs such as LLaMA \citep{touvron2023llama}, Falcon \citep{almazrouei2023falcon} and Gemma \citep{team2024gemma}. These open models offer the advantage that they can be used by end users for inference or local fine-tuning, provided their hardware has sufficient memory for the size of the models. However, ``full fine-tuning'' — a process that involves updating all previously trained parameters — is still prohibitively expensive for large models. For example, the standard 16-bits fine-tuning of the LLaMA-$65$B parameter model requires over $780$ GB of GPU memory \citep{dettmers2023qlora}. This high requirement is due to the need to store both the weights of the model and the states of the optimizer in GPU memory, a need that increases as the size of the LLMs increases.

A common method to mitigate memory constraints is to quantize the model weights, activations, and gradients — to a lower bit precision. Quantization-Aware Training (QAT) is often used in computer vision; see \cite{courbariaux2015binaryconnect, liu2020reactnet, gholami2022survey}. However, training large language models (LLMs) from scratch is impractical due to high computational cost. Post-training quantization (PTQ) is an efficient compromise \citep{dettmers2022gpt3, frantar2022gptq}, which has recently attracted much attention \citep{kim2023squeezellm, dettmers2023spqr, kim2023memory, shao2023omniquant}. Although most research focuses on scalar quantization (SQ), a few studies investigate LLM compression using vector quantization (VQ) \citep{tseng2024quip, egiazarian2024extreme}.

In \cite{dettmers2023qlora}, quantization is effectively combined with the Parameter Efficient Fine-Tuning (PEFT) method, LoRA \citep{hu2021lora}, to improve efficiency and practicality in memory-constrained environments. Post-Training Quantization (PTQ) has the potential to be further improved to achieve sub-$3$ bit quantization \citep{li2023loftq, guo2023lq}. However, it was found that the weights of the LLM often contain outliers — weights with significantly higher values than others \citep{kim2023squeezellm, dettmers2023spqr}. These outliers pose a considerable challenge for model compression with PTQ and lead to significant quantization errors.

\begin{figure}
    \centering
   \begin{subfigure}[b]{0.495\textwidth}
    \includegraphics[scale=0.35]{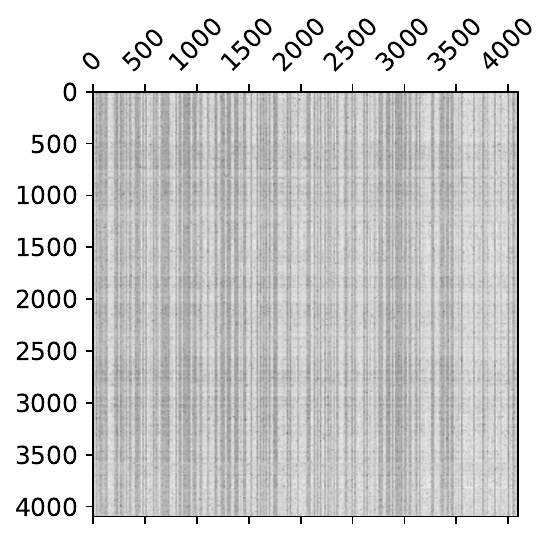}
    \includegraphics[scale=0.35]{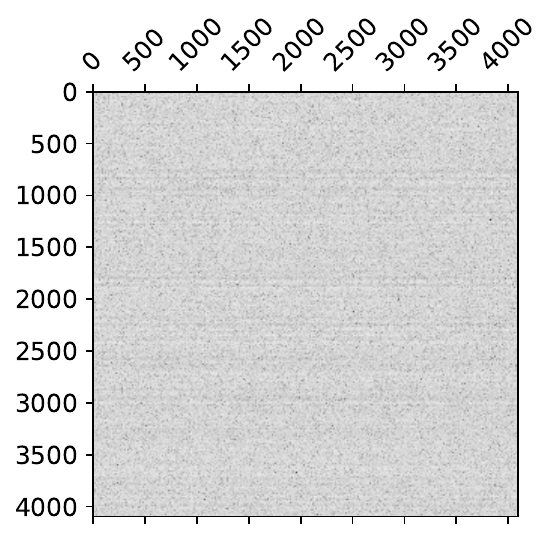}       \caption{Mistral-7B \citep{jiang2023mistral}}
   \end{subfigure}
   \hfill
   \begin{subfigure}[b]{0.495\textwidth}
    \includegraphics[scale=0.35]{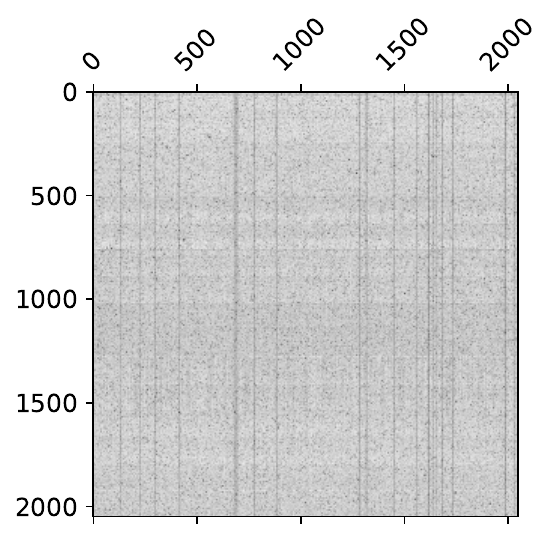}
    \includegraphics[scale=0.35]{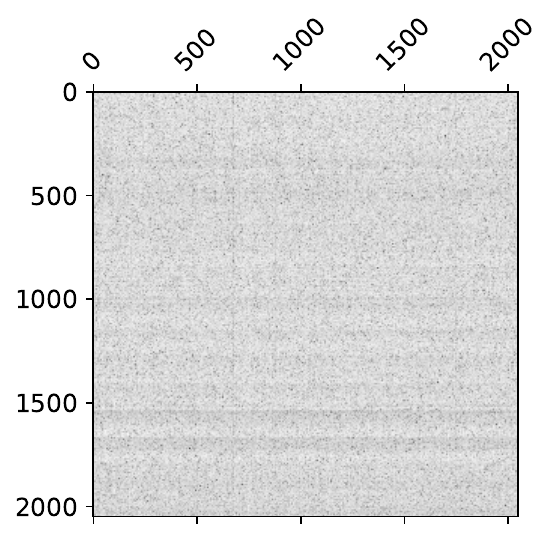}
    \caption{Gemma-2B \citep{team2024gemma}}
    \end{subfigure}
    \caption{Pre-trained matrix from the first block (left; with ``structures''), and pre-trained matrix from the last block (right) for two different models. Stronger vertical patterns appear in the first blocks.}
    \label{fig:structures}
\end{figure}

In this paper we present \Algo\ - for \textbf{Re}sidual \textbf{A}utoencoder \textbf{LLM} - a general approach for LLM PTQ and fine-tuning. Pre-trained LLM matrices are decomposed into a $16$-bit remainder (low rank, sparse outliers, etc.) and a compressed part, which is fed into a VQ autoencoder \citep{van2017neural}. In our experiments, we implement a low-rank and quantized decomposition of pre-trained LLM matrices. In this approach, only the low-rank components are fine-tuned (block-wise and end-to-end) while the quantized elements remain static. Our quantization strategy (\ie\ the shape of the autoencoder) adapts to the matrix patterns: Our results suggest that some pre-trained LLM matrices exhibit ``spatial'' patterns (see \Cref{fig:structures}; left) that bear similarities to those in images/videos and allow for highly effective compression (see \Cref{fig:error}).

\paragraph{Contributions:}
\begin{itemize}
\item We present \Algo, a method that uses a novel autoencoder and a residual pipeline to efficiently compress pre-trained LLM matrices;
\item We show that state-of-the-art PTQ approaches \citep{lin2023awq, shao2023omniquant, tseng2024quip, egiazarian2024extreme} and fine-tuning methods \citep{hu2021lora, dettmers2023qlora, guo2023lq, li2023loftq, liao2024apiq} are all special cases of \Algo;
\item We propose a preprocessing step that includes scaling and column permutations of matrices to mitigate the quantization errors associated with outliers; We also propose to adapt the general autoencoder scheme to the type of pre-trained matrix patterns.
\item Our approach demonstrates that fine-tuning end-to-end with block-wise error reduction leads to the best results reported in the literature for 3 and 2-bit Post-Training Quantization (PTQ).
\end{itemize}

\section{Related works}
\label{sec:related-works}
\paragraph{LLMs adapters.} After the introduction of high-performance open-source LLMs and due to the impracticality of ``full fine-tuning'', several methods of parameter-efficient fine-tuning (PEFT) have emerged, including prefix tuning \citep{li2021prefix}, selective fine-tuning \citep{guo2021parameter} and Low Rank Adapter (LoRA). LoRA, introduced in \cite{hu2021lora}, is a simple but effective fine-tuning method that retains the pre-trained matrices but adds a low-rank component. For a typical pre-trained matrix $W$ of size $4096 \times 4096$, LoRA introduces two additional matrices of size $4096 \times r$ and $r \times 4096$, where $r \ll 4096$, and tunes only their $2 \times r \times 4096$ parameters. In our work, we use DoRA \citep{liu2024dora} to further improve the fine-tuning by decomposing a weight into its magnitude and direction: $W_{\text{finetune}}= m \frac{W + L_1 (L_2)^t}{\Vert W + L_1 (L_2)^t\Vert_c}$, where $W$ is the frozen pre-trained weight, $m$ is the trainable size vector, $(L_1, L_2)$ are the low-rank (trainable) adapters, and $\Vert \cdot \Vert_c$ denotes the Euclidean norm of a matrix over each column. DoRA with the trainable size vector requires little computational effort, but can lead to significant performance improvements \citep{liu2024dora}.

\paragraph{Quantization.} Current methods for compressing LLMs predominantly use quantization techniques. Early strategies, such as ZeroQuant \citep{yao2022zeroquant} and nuQmm \citep{park2022lut}, relied primarily on direct rounding of weights to the nearest quantization level. Later developments improved this approach by handling outliers through quantization to higher bitwidths \citep{xiao2023smoothquant, dettmers2022gpt3, kim2023squeezellm, dettmers2023spqr}. Methods similar to \Algo\ include those that combine quantization with a low-rank decomposition; see \eg\ \cite{dettmers2023qlora, guo2023lq, li2023loftq, liao2024apiq}. QLoRA \citep{dettmers2023qlora} combined Parameter Efficient Fine-Tuning (PEFT) and quantization, but added zero-initialised low-rank adapters after quantization. In contrast, Loftq \citep{li2023loftq} and LQ-LoRA \citep{guo2023lq} propose to minimize quantization errors by initializing LoRA components with an SVD of the pre-trained weights. As part of this integration, ApiQ \citep{liao2024apiq} uses gradient descent to optimize both the LoRA components and the quantization parameters for the entire model rather than for each individual layer. Quantization of pre-trained weights facilitates efficient inference on devices with limited memory. To achieve significant computational and energy efficiency, recent studies have combined quantization of weights with activation quantization \citep{liu2023qllm,nrusimha2024mitigating}.

\paragraph{Block/Layer-Wise Tuning.} GPTQ \citep{frantar2022gptq} introduced a higher accuracy strategy using an approximate large-scale solver to minimize the layer-wise quadratic error, which is crucial for low bit-width quantization, as highlighted in \cite{tseng2024quip, egiazarian2024extreme}. Quip\# \citep{tseng2024quip} applies random rotations to the pre-trained matrices, segments the resulting matrix into vectors of dimension $d=8$ and uses optimal lattice quantizers \citep{viazovska2017sphere} to quantize each vector. Due to the random rotation, the distribution of the coefficient vector resembles an isotropic Gaussian distribution, but breaks the inherent dependence between the individual coefficients (see \Cref{fig:structures}). In contrast, AQLM \citep{egiazarian2024extreme} uses additive quantization with adaptive codebooks per layer and performs blockwise fine-tuning. Each codebook is first filled with Kmeans \citep{arthur2007k}, and the codewords are optimized to minimize the mean square error caused by the VQ at the output of each block. Quip\# and AQLM have achieved stable results (i.e. a single-digit increase in perplexity) in the compression range of $2$ bits per parameter.

\section{Method}
\label{sec:method}
\begin{figure}[ht]
    \centering
    \includegraphics[trim={0 9cm 0 0}, scale=0.5]{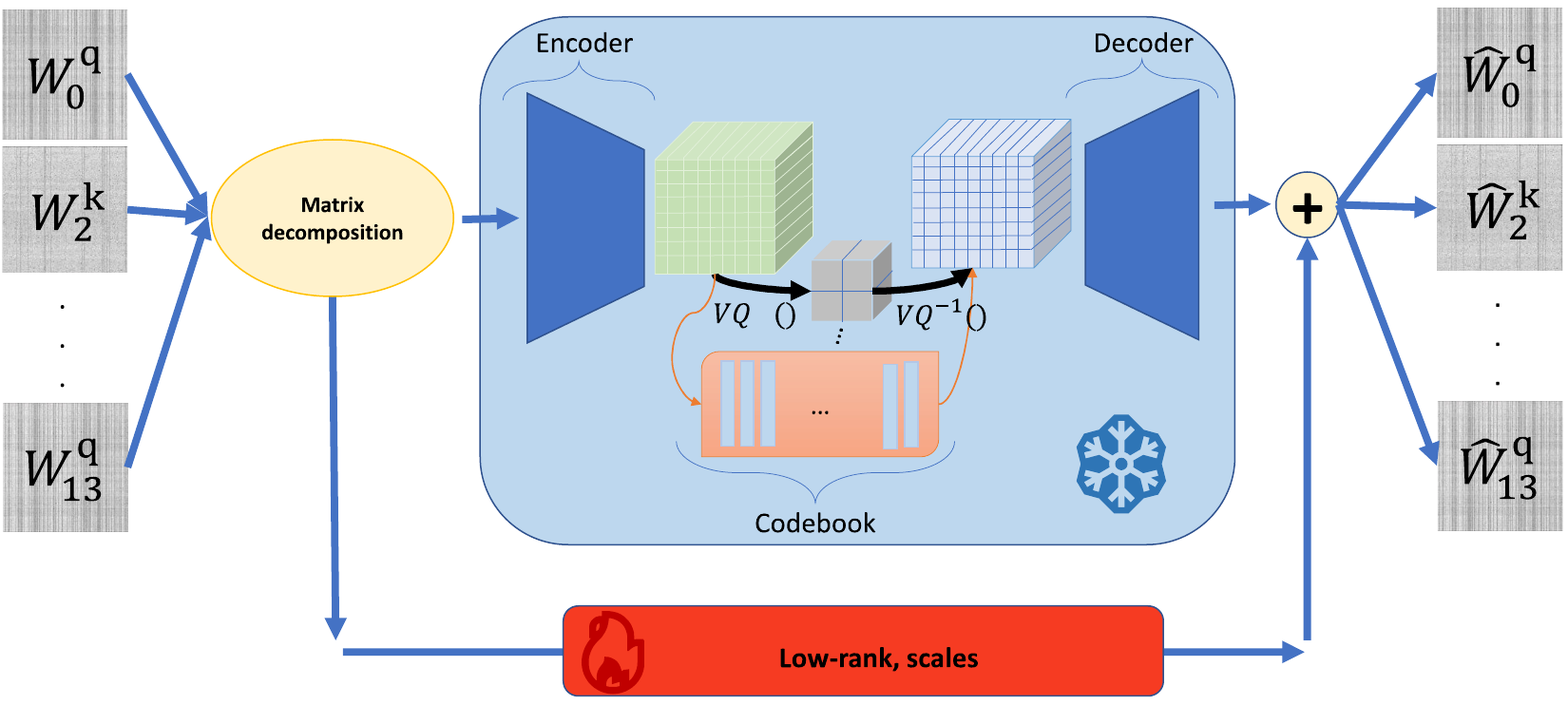}
    \caption{\Algo; during the fine-tuning step only low-rank and scales are updated}
    \label{fig:reallm}
\end{figure}

\paragraph{Low-rank/sparse decomposition.} Starting from a pre-trained LLM matrix $W \in \rset^{p\times q}$, $W$ is decomposed in a first step into a residual component $R \in \rset^{p\times q}$ and a quantized matrix $Q$ (which is represented on average with $b$ bits per coordinate). Only the residual matrix is retained with high bit accuracy and further optimized in the fine-tuning phase using a small calibration dataset. Any efficient matrix decomposition can fit into the residual part: butterfly \citep{dao2019learning}, sparse outliers \citep{dettmers2023spqr, lin2023awq}, etc. In \Cref{sec:experiments} we use a low-rank component \( R= L_1 (L_2)^t \). This structure is analogous to the \emph{data-free} method described in \cite{guo2023lq}. The aim is to identify \( Q \), \( L_1 \) and \( L_2 \) that (approximately) solve the following problem:
\begin{equation}
 \min_{Q, L_1, L_2} \Vert W - (Q + L_1(L_2)^t)\Vert.
\end{equation}
QLoRA \cite{dettmers2023qlora} provides a suboptimal solution for the previously described optimization problem by setting \( L_1 = 0 \) and solving \( \min_{Q} \| W - Q \| \). There is no guarantee that the initialization of the low-rank part to zero is optimal. It has been reported that QLoRA, Apiq and Loftq perform better than QLoRA in several language generation benchmarks \citep{guo2023lq, liao2024apiq, li2023loftq}.

\paragraph{Mixed-autoencoder configuration.} An autoencoder is the composition of an encoding function $\mathcal{E}$ and a decoding function $\mathcal{D}$. In \Algo, $\mathcal{E}_\psi$ and $\mathcal{D}_\phi$ are parameterized by neural networks $\psi, \phi$ and especially $\mathcal{E}_\psi: \rset^{p\times q} \xrightarrow{} \rset^{e_0 \times e_1 \times e_2}$, $\mathcal{D}_\phi: \rset^{e_0 \times e_1 \times e_2} \xrightarrow{} \rset^{p\times q}$, with $e_0e_1e_2 \ll pq$. As far as we know, most previous works on quantization of LLMs have focused on applying the same quantization strategy \emph{directly} to the (rotated) pre-trained matrix: \ie\ take the embedding dimensions $e_0=p, e_1=q, e_2=1$. Quip\# \citep{tseng2024quip} is a special case of \Algo\ (with no residual $R$) where the encoder is assumed to be a (random) rotation matrix $\mathcal{E}_\psi = U$ and the decoder is assumed to be the inverse $\mathcal{D}_\phi = U^{-1}$. LQ-LoRA \citep{guo2023lq}, Loftq \citep{li2023loftq}, and ApiQ \citep{liao2024apiq} are special cases of \Algo\ where the encoder and the decoder are defined as the identity matrix.

The approach may not be optimal as some matrices are more challenging to quantize than others  \citep{guo2023lq}. Specifically, \Cref{fig:structures} shows that pre-trained LLM matrices can display very different ``spatial'' patterns. \Algo\ adapts the autoencoder to the type and shape of the matrix. When quantizing pre-trained matrices with strong coefficient dependencies, \Algo\ is akin to image and video compression techniques that use the implicit neural representation \citep{chen2023hnerv, kwan2024hinerv}. \Algo\ extracts latent representations $\mathcal{E}_\psi(W)$ of a set of trained LLM matrices. In the next step, a decoder model is trained to generate the original LLM matrices $\mathcal{D}_\phi(\mathcal{E}_\psi(W))$ (refer to \Cref{fig:reallm}). During the inference phase of an LLM, only the latent embedding $\mathcal{E}_\psi(W)$ and the decoder $\mathcal{D}_\phi$ are needed to reconstruct the original weight $W$, with the exception of the additional low-rank and scale components. We use HNeRV \citep{chen2023hnerv} to train the autoencoder efficiently. HNeRV (over-)fits a model to the input matrices (\ie\ here the pre-trained LLM matrices) with an encoder $\mathcal{E}_\psi$ consisting of standard 2D convolutions, and a decoder combining 2D-convNeXt \citep{liu2022convnet} and PixelShuffle \citep{shi2016real}.

The decoding process is fast, as HNeRV requires only one network forward operation for decoding. \Algo\ compression is a combination of a small (\wrt\ input signals) neural decoder model $\mathcal{D}_\phi$ and model compression ($b_{\phi} \ll 16$). 
HNeRV implements weight pruning \citep{han2015learning}, weight quantization (PTQ) and entropy encoding. We go one step further by using a QAT approach: we train the decoder network $\mathcal{D}_\phi$ with convolution kernels quantized to $b_{\phi}=6$ bits during training with the straight-through estimator \citep{bengio2013estimating}. 
For a typical matrix of size $4096 \times 4096$, we train a decoder network with $c=7.2 \cdot 10^6$ parameters on $b_\phi=6$ bits and an embedding of size $16 \times 16 \times 16$. 
The total bit budget for the given matrix is therefore $\frac{6 \cdot (7.2 \cdot 10^6)+16 \cdot (16 \cdot 16 \cdot 16 \cdot \frac{4096^2}{512^2})}{(4096)^2}=2.82$ bits per coordinate.

\paragraph{Vector Quantization (VQ).} An efficient way to store the embedding $\mathcal{E}_\psi(W)$ with few bits is  VQ. AQLM \citep{egiazarian2024extreme} is a special case of \Algo\, where the latent representation is the matrix $W$ itself. AQLM optimizes multiple codebooks with gradient descent thanks to a calibration dataset. In contrast, for the forward pass, we opted for a \emph{data-free} vector quantization (VQ) method based on Kmeans \citep{arthur2007k}. A given embedding of size $e_0 \times e_1 \times e_2$ is divided into buckets of dimension $d$. First, we compute scales with NF-normalization \citep{dettmers2023qlora, guo2023lq}. The scales are further quantized following the idea of LQ-LoRA, resulting in an additional memory cost of $0.1$ bit \citep{guo2023lq}. Then we optimize $2^{b \cdot d}$ codewords using Kmeans clustering on the set of vectors in dimension $d$ to create a codebook. Each vector of dimension $d$ is quantized by the index of the closest element in the codebook (see \Cref{fig:reallm}). Consequently, the total number of bits required is $(b d)\frac{e_0e_1e_2}{d}$, \ie\ $b$ bits per coordinate. Additional memory is required to store the codebook (namely $16 \cdot d \times 2^{b \cdot d}$ bits). It should be noted that no separate gradient is defined for the quantization operator with the closest element \citep{van2017neural}. Therefore, during the backward pass, we approximate the gradient similarly to the straight-through estimator \citep{bengio2013estimating} and simply copy the gradients from the decoder input to the encoder output.

\begin{figure}
    \centering
   \begin{subfigure}[b]{0.495\textwidth}
  \includegraphics[scale=0.49]{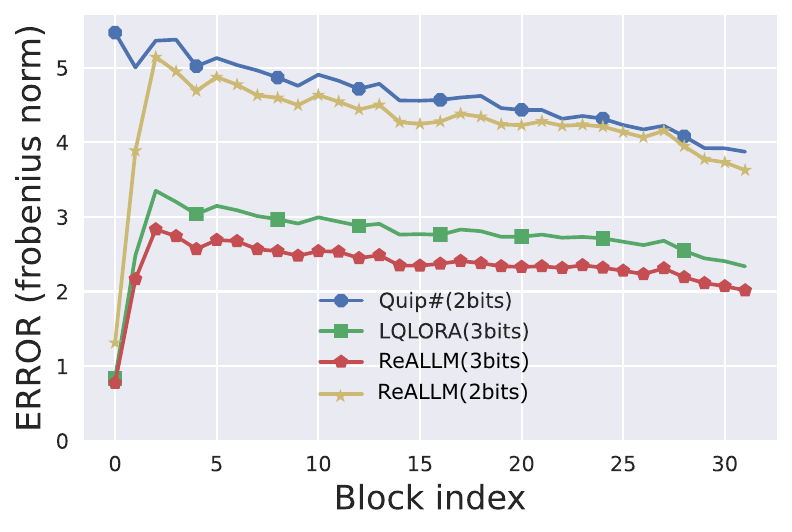}
     \caption{Mistral-7B \citep{jiang2023mistral}}
   \end{subfigure}
   \hfill
   \begin{subfigure}[b]{0.495\textwidth}
    \includegraphics[scale=0.49]{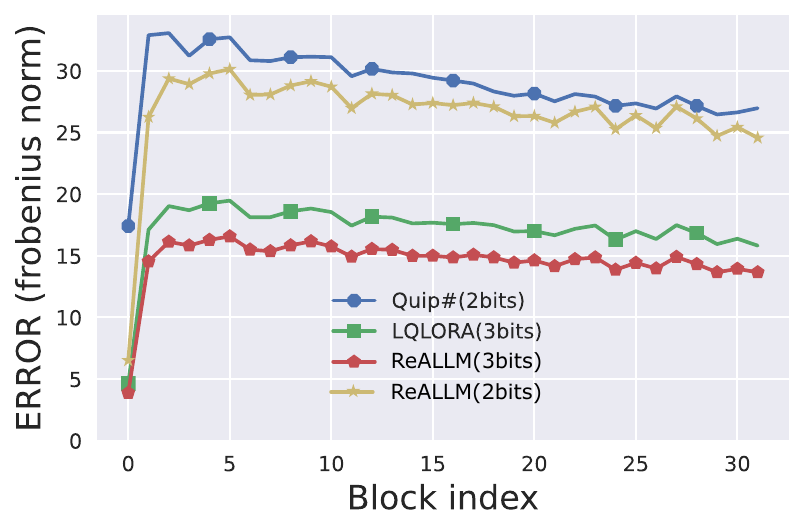}
     \caption{Llama2-7B \citep{touvron2023llama}}
    \end{subfigure}
    \caption{Reconstruction (Frobenius norm) error for layer of type ``Q'' for all blocks. Quip\# \citep{tseng2024quip} does not take advantage of the structures in the first blocks.}
    \label{fig:error}
\end{figure}

\paragraph{Quantization pre-processing.} Before using a tensor quantization method, it is important to perform an appropriate scaling. Several parameters (number of blocks, quantile bins, etc.) are chosen to correspond to a given compression ratio. But the presence of outliers \citep{kim2023squeezellm, dettmers2023spqr} forces the scaling and quantization methods to have a poor compression ratio \citep{lin2023awq, tseng2024quip, ashkboos2024quarot}. Incoherence processing uses random rotations as a pre-processing step. Although the main purpose of incoherence processing is to reduce the effects of outliers \citep{tseng2024quip, ashkboos2024quarot}, this technique has a detrimental effect on the structure of the pre-trained matrices within the initial blocks of the LLM (see \Cref{fig:structures,fig:error}). This is a serious bottleneck as quantization errors in these initial blocks can propagate throughout the model. As shown in \Cref{fig:structures}, some matrices have no specific patterns and resemble random Gaussian noise interspersed with randomly positioned outliers. To deal with outliers in the latent representation, we suggest rearranging the columns to create some spatial regularity. This strategy aims to find the most effective permutations that cluster outliers. \cite{trukhanov2024accurate} has recently elaborated a row/column permutation strategy that summarizes vectors (\ie\ sets of rows or columns) with similar norms. In contrast, for \Algo\ we propose to permute columns such that neighboring columns are ``similar'' and not just on the same hypersphere. We develop a basic, yet efficient method for this: first we select a block of size $128 \times q$ in the input tensor of size $e_0 \times e_1 \times e_2$. We start from the first vector, and we search for its closest neighbor in the set of $(q-1)$ vectors (we compute $(q-1)$ scalar products and select the vector that minimizes it). Then, we permute the neighbor vector with the vector in the second position of the block. The process is then iterated; more details are given in \Cref{algo:permute} and \Cref{sec:supp_permutations}. Note that the memory storage of the permutation is negligible: for a LLM matrix with $q=4096$ columns, the permutation requires $q \log(q) = 12 \times 4096$ additional bits for each block of size $128 \times 4096$, hence the memory overhead is about $0.09$ bits per coordinate.
\begin{algorithm}[ht]
\caption{permutation function\label{algo:permute}}
\SetKwInOut{Input}{Input}
\SetKwInOut{Output}{Output}
\SetKwBlock{Loop}{Loop}{end}
\SetKwBlock{Initialize}{Initialize}{end}
\SetKwFor{When}{When}{do}{end}
\SetKwFunction{Wait}{Wait}
\SetKwFunction{ClientMain}{ClientLocalTraining}
\SetAlgoLined
\Input{Matrix $w$ of size $128 \times q$ \;}
\For{$j=0,\dots,q-1$}{
$column_{j} = w[:, j]$ \;
$indx_{j} = get\_index\_nn(column_{j}, w[:,j+1:q])$ \tcc{get the nearest neighbor index of current $column_{j}$, among the rest of un-permuted columns $w[:, j+1:q]$}
Permute $w[:,j+1]$ and $w[:,indx_{j}]$\;
Save the inverse of the permutation index in $inv\_permut$ \;
}
\Output{$w, inv\_permut$}
\end{algorithm}

\paragraph{\Algo: a new LLM format.} LLM standard formats represent LLM weights as a set of matrices encoded on $16$ bits. Scalar quantization approaches \citep{frantar2022gptq, dettmers2023qlora} represent any matrix of size $p \times q$ with $b \cdot pq$ bits for a budget of $b$ bits. Vector quantization (VQ) methods \citep{egiazarian2024extreme, tseng2024quip} represent any matrix of size $p \times q$ with a smaller matrix of size $p \times \frac{q}{d}$ with $b \cdot d$ bits for a budget of $b$ bits and a vector dimension $d$. \Algo\ goes one step further and proposes to represent each matrix of size $p\times q$ with a small embedding of size $e_0 \times e_1 \times e_2$ on $b$ bits and a neural decoder model $\mathcal{D}_\phi$ with $c$ parameters on $b_{\phi}$ bits. \Cref{fig:reallm} illustrates the most important innovation of \Algo: LLMs are no longer represented by a set of matrices, but as a combination of embeddings and a single neural decoder model. \Algo\ learns a single model for a specific family of basic models (\eg\ LLaMAs, Gemma). If a specific weight matrix is needed for a specific LLM, one must take its embedding and perform only one \emph{single} forward pass with the decoder $\mathcal{D}_\phi$. This speeds up the decoding step compared to diffusion-based approaches \citep{wang2024neural, soro2024diffusion}. Note that for \Algo\ a decoder model has to be trained on LLM matrices, but this learning step is done once and for all. Additionally, the more we train and overfit, the better \Algo\ becomes.

The set of hyper-parameters for \Algo\ are: $r$ the rank, $(e_0, e_1, e_2)$ the shape of the latent representation, $(b,d)$ the number of bits and the bucket dimension in the VQ, and $(c, b_\phi)$ the number of parameters and the number of bits of the decoder. We have conducted extensive experiments to find suitable configurations; however, we were unable to test configurations with a large decoder size. For \eg, for small embeddings ($e_0e_1e_2<1024$) and a total budget of $3$ bits for a single LLaMA2-7B model, the decoder model in \Algo\ has $c=3.5 \cdot 10^9$ parameters trained on $b_{\phi}=6$ bits. Our GPU is unable to accommodate multiple LLM matrices in memory for \Algo\ training, typically with size $n \times n;n>4096$. Therefore, we test \Algo\ on a set of $512 \times 512$ patches extracted from pre-trained LLM matrices, and we use the HNeRV \cite{chen2023hnerv} autoencoder model. For more details on the practical aspect of decoder training, see \Cref{sec:supp_limitations}.

We have experimentally discovered two sets of optimal combinations of hyperparameters that depend on the type and shape of the pre-trained matrix. Some pre-trained matrices, especially those closer to the input tokens, compress better with small latent representations ($e_0e_1e_2<1024$) in high bit precision ($b >8$) and (relatively) large decoders ($c > 4 \cdot 10^6$). Other pre-trained matrices (usually deeper in the LLM) compress better with very large embeddings ($e_0>\frac{p}{4}, e_1 > \frac{q}{4}, e_2 \in [1, 2]$) with low bit precision ($b \ll 8$) and (relatively) small decoders ($c \ll 10^6$).
In \Cref{fig:error} \Algo\ achieves the lowest Frobenius norm quantization error. We perform ablation experiments with this metric to decouple the effects of VQ and permutation preprocessing of \Algo\ on the final performance. For example, in block $8$ (Mistral-7b; left panel), the error for scalar quantization (SQ; used in \cite{dettmers2023qlora, guo2023lq}) is $2.96$. This error decreases with VQ to $2.68$ and with permutation further to $2.54$, while permutation alone (\ie\ with SQ) leads to an error of $2.88$. Quip\# rotates the matrices randomly, causing all patterns in the initial blocks to be lost.

\begin{algorithm}[ht]
\caption{Pseudo-code for \Algo\ with block-wise and end-to-end fine-tuning \label{algo:general_method}}
\SetKwInOut{Input}{Input}
\SetKwInOut{Output}{Output}
\SetKwBlock{Loop}{Loop}{end}
\SetKwBlock{Initialize}{Initialize}{end}
\SetKwFor{When}{When}{do}{end}
\SetKwFunction{Wait}{Wait}
\SetKwFunction{ClientMain}{ClientLocalTraining}
\SetAlgoLined
\Input{Number of end-to-end fine-tuning steps $\nsteps$, Number of block-wise fine-tuning steps $K$, Number of blocks $n$, Shape of the latent space ($e_0, e_1, e_2$), Number of weights in the decoder $c$, Number of bits for the decoder weights $b_\phi$, Number of VQ bits per dimension $b$, VQ dimension $d$, Rank $r$ \;}
\Initialize{
Get pre-trained matrices $\{W^q, W^k, W^v, W^o, W^{gate}, W^{up}, W^{down}\}$ for all $n$ blocks \;
}
\tcc{\textbf{Block-wise fine-tuning}}
\For{$j=0,\dots,n-1$}{
$B_j = \{W^q, W^k, W^v, W^o, W^{gate}, W^{up}, W^{down}\}[block=j]$\;
$output_j = forward\_pass(B_j)$ \tcc{get \emph{non}-quantized output}
\For{$l \in \{q, k, v, o, gate, up, down\}$}{

$L1^l_j, L2^l_j = svd\_decomposition(W^l_j, rank=r)$ \;
$W^l_j = W^l_j - L1^l_j (L2^l_j)^t$ \;
${\mathcal{E}_\psi}(W^l_j), {\mathcal{D}_\phi}^l_j = autoencoder(W^l_j, e_0, e_1, e_2, c, b_\phi)$ \tcc{latent representation and decoder}
${\mathcal{E}_\psi}(W^l_j), inv\_permut^l_j = permute({\mathcal{E}_\psi}(W^l_j))$ \tcc{with \Cref{algo:permute}}
${\mathcal{E}_\psi}(W^l_j) = normalize({\mathcal{E}_\psi}(W^l_j))$ \tcc{with NF-normalization \citep{dettmers2023qlora, guo2023lq}}
$codebook^l_j = Kmeans({\mathcal{E}_\psi}(W^l_j), b, d)$\;
$codes^l_j = get\_index\_nn({\mathcal{E}_\psi}(W^l_j), codebook^l_j)$ \tcc{get nearest neighbor index in $codebook^l_j$}
\textcolor{magenta}{$W^l_j \gets \{codes^l_j, codebook^l_j, {\mathcal{D}_\phi}^l_j, inv\_permut^l_j, L1^l_j, L2^l_j \}$}\;

$dora^l_j = DoRA(W^l_j, L1^l_j, L2^l_j)$ \tcc{get DoRA scale}

}
$dora\_quantized\_output_j = forward\_pass\_quantized(\{dora^l_j, L1^l_j, L2^l_j, W^l_j\}_{l \geq 0})$ \tcc{get output after quantization and DoRA}
$L_j = \Vert output_j - dora\_quantized\_output_j \Vert^2 $ \;
\For{$k=0,\dots,K-1$}{
\textcolor{red!75}{Optimize $\{dora^l_j, L1^l_j, L2^l_j\}_{l \geq 0}$} with gradient descent to minimize $L_j$\;
}
}
\tcc{\textbf{End-to-end fine-tuning}}
\For{$t=0,\dots,\nsteps-1$}{
\textcolor{red!100}{Optimize $\{dora^l_j, L1^l_j, L2^l_j\}_{l,j \geq 0}$} with gradient descent \;
}
\end{algorithm}

\section{Experimental validation}
\label{sec:experiments}
We test \Algo\ on the LLaMA-2 \citep{touvron2023llama} family models (with $7$ and $13$ billions parameters). We compare our method with other quantization approaches for a budget of $3$ and $2$ bits per coordinate. We partially reused code from the implementations of LQ-LoRA\footnote{https://github.com/HanGuo97/lq-lora/tree/main}, AQLM \footnote{https://github.com/Vahe1994/AQLM} and HNeRV\footnote{https://github.com/haochen-rye/HNeRV}. On an Nvidia A40 GPU (with $46$GB memory), the entire computation (PTQ + fine tuning) takes about $90$ hours for a LLaMA2-7B model.

\paragraph{Language Generation Tasks.} For continual language modeling, we train on a single partition of the C4 \citep{raffel2020exploring} dataset for half an epoch and use a sequence length of $4096$ for training only. Note that the native context length for LLaMA-2 \citep{touvron2023llama} is $4096$, while it is $2048$ for LLaMA-1. Consequently, in the literature LLaMA-2 models are evaluated with token sequences of size $2048$ (all except \citep{egiazarian2024extreme} follow this rule). Therefore, we use a sequence length of size $2048$ for both WikiText-2 \citep{merity2016pointer} and C4 evaluation.

Our main baselines are LQ-LoRA \citep{guo2023lq}, Quip\# \citep{tseng2024quip}, and AQLM \citep{egiazarian2024extreme}. However, we also report the performance of popular quantization approaches GPTQ \citep{frantar2022gptq}, AWQ \citep{lin2023awq}, Omniquant \citep{shao2023omniquant}, as well as the performance of recent work ApiQ \citep{liao2024apiq} and QuaRot \citep{ashkboos2024quarot}. In the results below, we present the target bits per parameter that takes into account quantized weights and include parameters kept in high precision (head layer, scales, codebooks, permutations in $16$ bits, and low-rank matrices in $8$ bits precision) similarly to the related work. The exact bit budget is detailed in \Cref{tab:representation} in the Appendix.

In our experiments, following \cite{dettmers2023qlora, guo2023lq}, we take a DoRA \citep{liu2024dora} rank of $r=64$ (unless otherwise specified), we set the decoder bit precision to $b_\phi=6$, and we adjust the size of the latent representation $(e_0, e_1, e_2)$ depending on the block index (tested from $(4096, 4096, 1)$ to $(16, 16, 16)$), and we have tested several VQ in dimension $d=2$ or $d=4$. The VQ-autoencoder is trained with cosine scheduler with a maximum learning rate of $0.001$ for $2000$ epochs. Then we (optionally) tune the low-rank components block-wise with a batch of size $32$ and a step size of $1\cdot e^{-5}$. The end-to-end fine-tuning is run with batches of size $1$, and a learning rate of $2\cdot e^{-5}$. As far as we know, we have also developed the first VQ code (available in the supplementary material) that makes efficient use of PyTorch’s ``torch dispatch'' functionality \citep{Ansel_PyTorch_2_Faster_2024}, which is known to be as fast as dedicated CUDA kernels \citep{guo2023lq}. This allows us to overload PyTorch operations to perform just-in-time dequantization.

In \Cref{tab:perplexity_7b_ablation,tab:perplexity_7b} we evaluate the perplexity of \Algo\ on the respective validation datasets of C4 and WikiText-2 for a single run. During fine-tuning (on a single partition of the C4 dataset), we only update the DoRA components (scales and low-rank matrices). For each dataset, we provide three sets of results in \Cref{tab:perplexity_7b_ablation}: Perplexity without any fine-tuning (only low-rank and VQ autoencoder decomposition), perplexity with only block-wise fine-tuning, and perplexities with end-to-end fine-tuning (in addition to the block-wise fine-tuning process).
\begin{table}[ht]
  \caption{Perplexity $(\downarrow)$ on the validation dataset for LLaMA2-7B, with a sequence length of $2048$}
  \label{tab:perplexity_7b_ablation}
  \centering
  \begin{tabular}{cccc|c|c}
  \toprule
    Method & \#bits & rank $r$ & bucket $d$ & C4 $(\downarrow)$ & WikiText-2 $(\downarrow)$\\
    \midrule
      \Algo\ (no fine-tuning) & $3$ & $32$ & $2$ & $7.78$ & $6.21$ \\
      \Algo\ (block-wise) & $3$ & $32$ & $2$ & $7.56$ & $6.01$ \\
      \Algo\ (40\% training) & $3$ & $32$ & $2$ & $7.31$ & $5.80$ \\
    \Algo\ (full training) & $3$ & $32$ & $2$ & $7.29$ & $5.79$ \\
    \midrule
   \Algo\ (no fine-tuning) & $3$ & $64$ & $2$ & $7.72$ & $6.10$ \\
   \Algo\ (block-wise) & $3$ & $64$ & $2$ &  $7.51$ & $5.92$ \\
     \Algo\ (40\% training) & $3$ & $64$ & $2$ & $7.30$ & $5.78$ \\
   \Algo\ (full training) & $3$ & $64$ & $2$ & $7.27$ & $5.77$ \\
    \midrule
    \Algo\ (no fine-tuning) & $2$ & $64$ & $2$ &  $45.96$ & $51.74$ \\
    \Algo\ (block-wise 50 epochs) & $2$ & $64$ & $2$ &  $18.61$ & $16.95$ \\
    \Algo\ (block-wise 200 epochs) & $2$ & $64$ & $2$ &  $10.11$ & $8.31$ \\
    \Algo\ (40\% training) & $2$ & $64$ & $2$ & $8.56$ & $6.95$ \\
    \Algo\ (full training) & $2$ & $64$ & $2$ & $8.47$ & $6.91$ \\
    \midrule
    \Algo\ (no fine-tuning) & $2$ & $64$ & $4$ & $41.02$ & $40.85$ \\
    \Algo\ (block-wise 50 epochs) & $2$ & $64$ & $4$ & 15.74 & 12.08 \\
    \Algo\ (40\% training) & $2$ & $64$ & $4$ & $8.36$ & $6.74$ \\
    \Algo\ (full training) & $2$ & $64$ & $4$ & $8.28$ & $6.69$ \\

    \bottomrule
  \end{tabular}
\end{table}
Our \emph{data-free} version of \Algo\ (no fine-tuning; see \Cref{tab:perplexity_7b_ablation}) achieves state-of-the-art metrics for $3$ bit quantization. However, for a budget of $2$ bits, quantization errors are larger, and our results show that fine-tuning (both block-wise and end-to-end) is needed to further improve performance. This result is in line with the PTQ literature \citep{frantar2022gptq, egiazarian2024extreme}. \Cref{tab:perplexity_7b_ablation} also shows that reducing the rank from $r=64$ to $r=32$ has minimal effect on the final perplexity result, while halving the number of parameters that need to be tuned.  Moreover, a larger VQ dimension $d=4$ instead of $d=2$ leads to better results. Note that increasing $d$ comes at an additional storage cost (as explained in \Cref{sec:method}, $16 \cdot d \times 2^{b \cdot d}$ bits are needed to store the codebook).  Additional results for other models are available in the Appendix.
\begin{table}[ht]
  \caption{Perplexity $(\downarrow)$ on the validation dataset for LLaMA2-7B and LLaMA2-13B, with a sequence length of $2048$}
  \label{tab:perplexity_7b}
  \centering
  \begin{tabular}{cc|cc|cc}
  \toprule
    Method & Number of bits & \multicolumn{2}{c|}{C4 $(\downarrow)$} & \multicolumn{2}{c}{WikiText-2 $(\downarrow)$} \\
    & & 7B & 13B & 7B & 13B \\
    \midrule
    LLaMA2 \citep{touvron2023llama} & $16$ & $6.97$ & $6.46$ & $5.47$ & $4.48$\\
    \midrule
    GPTQ \citep{frantar2022gptq} & $3$ & $7.89$ & $7.00$ & $6.29$ & $5.42$ \\
    AWQ \citep{lin2023awq} & $3$ & $7.84$ & $6.94$ & $6.24$ & $5.32$ \\
    Omniquant \citep{shao2023omniquant} & $3$ & $7.75$ & $6.98$ & $6.03$ & $5.28$\\
    LQ-LoRA \citep{guo2023lq} & $3$ & $7.88$ & $-$ & $6.48$ & $-$ \\
    LoftQ \citep{li2023loftq} & $3$ & $-$ & $-$ & $5.63$ & $5.13$\\
    ApiQ[PTQ] \citep{liao2024apiq} & $3$ & $7.84$ & $6.88$ & $6.19$ & $5.18$ \\
    Quip\# \citep{tseng2024quip} & $3$ & $7.32$ & $6.72$ & $\textbf{5.79}$ & $5.10$\\
    QuaRot[A16W3] \citep{ashkboos2024quarot} & $3$ & $-$ & $-$ & $6.09$ & $5.37$\\
    \Algo\ & $3$ & \textbf{7.27} & $\textbf{6.69}$ & \textbf{5.77} & $\textbf{5.14}$ \\
    \midrule
    LoftQ \citep{li2023loftq} & $2$ & $-$ & $-$ & $7.85$ & $7.69$ \\
    ApiQ \citep{liao2024apiq} & $2$ & $-$ & $-$ & $7.46$ & $6.29$\\
    Quip\# \citep{tseng2024quip} & $2$ & $8.35$ & $\textbf{7.45}$ & $6.66$ & $5.74$\\
    AQLM \citep{egiazarian2024extreme} & $2$ & $8.56$ & $7.51$ & $\textbf{6.64}$ & $\textbf{5.65}$\\
    \Algo\ & $2$ & \textbf{8.28} & $\textbf{7.50}$ & \textbf{6.69}& $\textbf{5.72}$ \\
    \bottomrule
  \end{tabular}
\end{table}
In \Cref{tab:perplexity_7b} we compare \Algo\ with end-to-end fine-tuning, and the best performing PTQ approaches. All the methods cited in \Cref{tab:perplexity_7b} also uses a calibration dataset. It is interesting to note that \Algo\ with $2$ bits bridges the gap with the famous GPTQ \citep{frantar2022gptq} method on $3$ bits for the LLaMA2-13B. One major difference between \Algo\ and Quip\# \citep{tseng2024quip} is that the quantized weights are kept frozen during all the fine-tuning process in \Algo. As a consequence, we can store a single version of the quantized weight, and fine-tune several versions of the learnable parameters (\ie\ DoRA scales and low-rank matrices) for several fine-tuning tasks. On the contrary Quip\# updates all the weights (in $16$ bits precision) during the layer-wise fine-tuning. This does not only slow down the PTQ process (as gradients must be store for all weights in the given block), but it also means Quip\# has to store learnable vectors and also quantized weights for \emph{each} fine-tuning task.

\begin{table}[ht]
    \caption{Accuracy $(\uparrow)$ in LM Eval (acc, not acc\_norm).}
    \label{tab:fewshot}
    \centering
    \resizebox{\linewidth}{!}{
    \begin{tabular}{ccc|cccc|c}
    \toprule
       Method & Size & \#bits & ARC-challenge & ARC-easy & PiQA & Winogrande & Average\\
       \midrule
       LLaMA-2 & 7B & $16$ & $43.52$ & $76.26$ & $78.07$ & $69.22$ & $66.77$\\
       AQLM \citep{egiazarian2024extreme} & 7B & $2$ & $33.55$ & $62.79$ & $73.54$ & $64.61$ & $58.62$\\
       Quip\# \citep{tseng2024quip} & 7B & $2$ & $34.63$ & $64.60$ & $75.12$ & $64.89$ & $59.81$\\
       \Algo\ & 7B & $2$ & $\textbf{35.15}$ & $\textbf{68.56}$ & $\textbf{75.73}$ & $\textbf{66.46}$ & $\textbf{61.47}$\\
       \midrule
       LLaMA-2 & 13B & $16$ & $48.32$ & $78.48$ & $80.01$ & $72.13$ & $69.74$\\
       AQLM \citep{egiazarian2024extreme} & 13B & $3$ & $43.63$ & $73.51$ & $77.78$ & $67.56$ & $65.62$\\
       Quip\# \citep{tseng2024quip} & 13B & $3$ & $44.02$ & $72.45$ & $78.40$ & $69.13$ & $66.00$\\
       \Algo\ & 13B & $3$ & $\textbf{47.01}$ & $\textbf{75.96}$ & $\textbf{78.67}$ & $\textbf{70.96}$ & $\textbf{68.15}$\\
    \bottomrule
    \end{tabular}
    }
\end{table}
\paragraph{Zero-Shot Tasks.} Following HuggingFace’s Open LLM Leaderboard\footnote{https://huggingface.co/spaces/HuggingFaceH4/open\_llm\_leaderboard}, and the literature \citep{frantar2022gptq, guo2023lq}, we also measure zero-shot accuracy on ARC \citep{clark2018think}, PiQA \citep{tata2003piqa}, and Winogrande \citep{sakaguchi2021winogrande}, via the LM Evalaluation Harness \citep{gao2021framework}. We report results in \Cref{tab:fewshot}, and compute the average on the 4 mentioned tasks. For all LLM sizes, \Algo\ provides a notable advantage (between $0.5$ and $3$ points of accuracy improvement) with respect to AQLM \citep{egiazarian2024extreme} and Quip\# \citep{tseng2024quip}. Interestingly, the LLaMA2-13B model compressed on $3$ bits with \Algo\ performs better than the standard LLaMA-2-7B model ($16$ bits) on the zero-shot tasks.

\section{Conclusion}
We present \Algo, a weight-only PTQ method that achieves state-of-the-art results on LLMs at $2$, and $3$ bits budget. Our (low-rank) fine-tuning approach enables one to fine-tune language models with $13$ billions parameters on a \emph{single} GPU with less than $40$ GB of RAM.

Large context sequence lengths result in large $KV$-cache memory consumption during inference, and PTQ is a promising approach for compressing $KV$-cache activations \citep{hooper2024kvquant, ashkboos2024quarot}. Concurrently to our work, \cite{trukhanov2024accurate} propose a quantization method based on permutations of rows from $K$ and $V$ matrices. We are currently studying how to adapt \Algo\ to $KV$-cache quantization, and how to combine it with activation quantization.

\section{Societal impact}
This paper presents work whose goal is to advance the field of LLM compression and fine-tuning. There are many potential societal consequences of our work, in particular malicious usage of LLMs for spams or language generation on edge devices. However, this negative societal impact is not limited to \Algo, but to the field of LLM in general.

\clearpage
\newpage
\bibliographystyle{apalike}
\bibliography{biblio}


\clearpage
\newpage
\appendix

\section{Appendix / supplemental material}
\subsection{Structures in pre-trained matrices}
\label{sec:supp_structures}
\begin{figure}[ht]
    \centering
    \includegraphics[scale=0.7]{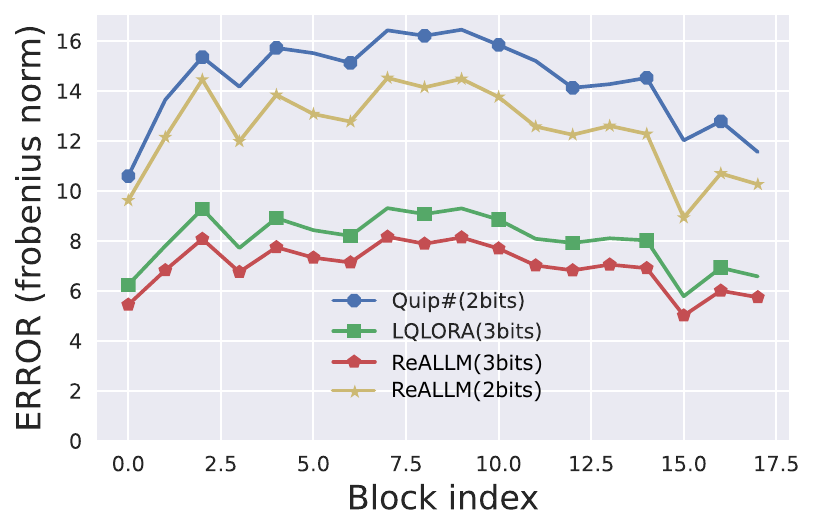}
    \caption{Reconstruction (Frobenius norm) error for layer of type ``Q'' for all blocks of Gemma2b LLM.}
    \label{fig:error_gemma2b}
\end{figure}

Interestingly, the blocks that show some visual structures in LLaMA and Mistral models are not the same for Gemma LLMs. For instance in \Cref{fig:error_gemma2b}, we can see that Gemma2b \citep{team2024gemma}'s matrices keep some internal patterns in all blocks, not only at the very first blocks. Note this has no negative impact on \Algo, as the shape of the encoder is experimentally adapted to each block.

\subsection{Autoencoder computational limitations}
\label{sec:supp_limitations}
Our GPU can not directly work on LLM pre-trained matrices with large sizes (typically of shape $4096 \times 4096$). Instead, we choose to split each pre-trained matrix into a set of $64$ ``patches'' of shapes $512 \times 512$, and we learn the decoder on the set of matches rather than on the big initial matrix. During the inference time, when de-quantizing a LLM matrix, we reconstruct each patch (in parallel) and we concatenate the patches together. This step of concatenation has a minimal impact on the final time complexity of our method.
In \Cref{tab:hnerv_full}, we present ablation experiment results on the type of decoder weight (only) quantization. We performed a quantization aware training approach, \ie\ directly optimizing weight quantized on $b_\phi$ bits using straight through estimator \cite{bengio2013estimating}. We also tested a post training quantization method where the weight of the decoder are quantized with a round to nearest (RTN) approache, at the end of the decoder training steps.
\begin{table}[ht]
    \centering
          \captionof{table}{Reconstruction (Frobenius norm) error for layer of type ``Q'' inside the first block of Mistral-7b model, for patches of size $512 \times 512$ using a constant embedding size of $(e_0, e_1, e_2)=(16, 16, 16)$, and a varying quantization strategy (during the decoder training, \ie\ QAT, or after the training, \ie\ PTQ).}
    \label{tab:hnerv_full}
    \centering
    \begin{tabular}{c|cccc}
\toprule
Error & \# parameters c ($\times 10^6$) & $b_\phi$ & bit budget & quantization \\
\midrule
$0.84$ & $-$ & $-$ & $3$ & NF3\citep{guo2023lq} \\
\midrule
$1.78$ & $7.2$ & $6$ & $2.82$ & PTQ \\
$1.19$ & $5.4$ & $7$ & $2.44$ & PTQ \\
$1.61$ & $7.7$ & $5$ & $2.32$ & QAT \\
$1.24$ & $4.5$ & $8$ & $2.21$ & QAT \\
$0.69$ & $7.2$ & $6$ & $2.82$ & QAT
    \end{tabular}
\end{table}
We vary the number of parameters $c$, and the bit precision $b_{\phi}$ of the decoder to target a total bit cost below $3$ bits per coordinate. This experiment show two different results: first, the influence on the quantization performance of the number of decoder parameters $c$ and their respective bit precision $b_\phi$ is not straightforward. Second, under the same parameters (number of parameters and bits), QAT gives better performance than the respective PTQ approach. Furthermore, for a reduced number of bits ($2.82$ vs $3$), \Algo\ yields a smaller quantization error compared to the scalar quantization NF3 \citep{dettmers2023qlora, guo2023lq}.

\begin{table}[ht]
    \caption{Comparison of several LLM format for $m$ matrices of size $p \times q$, and a budget of $b$ bits per coordinate. \Algo\ uses a decoder model with $c$ parameters trained on $b_{\phi}$ bits, and a rank $r$.}
    \label{tab:representation}
    \centering
    \resizebox{\linewidth}{!}{
    \begin{tabular}{c|ccc}
    \toprule
        Method & LoRA & VQ only (like AQLM) & \Algo\ \\
        \midrule
        Matrix representation & $(p \times q) \cdot 16$ & $(p \times \frac{q}{d}) \cdot b \cdot d$ & $(e_0 \times \frac{e_1}{d} \times e_2) \cdot b \cdot d$ \\
        Codebook & $-$ & $2^{bd} \cdot d \cdot 16$ & $2^{bd}\cdot d\cdot 16$ \\
        Decoder & $-$ & $-$ & $cb_{\phi}$ \\
        Low-rank & $(2 \times r \times \min(p,q))\cdot 16$ & $-$ & $(2 \times r \times \min(p,q))\cdot 16$ \\
        Total bit cost & $16(pq + 2r\min(p,q)) \cdot m$ & $(bpq + 2^{bd+4}d) \cdot m$ & $cb_\phi + 32r \min(p,q)+m(16d2^{bd}+e_0e_1e_2b)$ \\
    \end{tabular}
    }
\end{table}

\begin{table}[ht]
    \caption{Quantization and fine-tuning approaches as particular case of \Algo\ (with a rank $r$, and a budget of $b$ bits for VQ in dimension $d$) for a matrix of size $p \times q$.}
    \label{tab:as_reallm}
    \centering
    \resizebox{\linewidth}{!}{
    \begin{tabular}{c|ccccc}
    \toprule
        Method & rank $r$ & Autoencoder &  Latent $(e_0, e_1, e_2)$ & VQ dim. $(d)$ & VQ bits $(b)$ \\
        \midrule
        LoRA \citep{hu2021lora} & $64$ & None & $(p, q, 1)$ & $1$ & $16$ \\ 
        GPTQ \citep{frantar2022gptq} & $0$ & None & $(p, q, 1)$ & $1$ & $4$ \\ 
        QLoRA \citep{dettmers2023qlora} & $64$ & None & $(p, q, 1)$ & $1$ & $4$ \\ 
        LQ-LoRA \citep{guo2023lq} & $64$ & None & $(p, q, 1)$ & $1$ & $3$ \\ 
        Quip\# \citep{tseng2024quip} & $0$ & Rotation matrix & $(p, q, 1)$ & $8$ & $2$ \\ 
        AQLM \citep{egiazarian2024extreme} & $0$ & None & $(p, q, 1)$ & $8$ & $2$ \\
        \midrule
        \Algo\ & $64$ & Trainable & $(e_0, e_1, e_2)$ & $4$ & $2$      
    \end{tabular}
    }
\end{table}

\subsection{Permutations}
\label{sec:supp_permutations}
In \Algo, we compute permutations on sets of vectors in dimension $128$. We could work with smaller blocks, but it induces more memory dedicated to the permutation storage (one permutation for each block).

We start from the first vector (\ie\ the first column of the initial matrix shrunk to a dimension $d=128$), and we search for its closest neighbor in the set of $(q-1)$ vectors (we compute $(q-1)$ scalar products and select the vector that minimizes it). Then, we permute the neighbor vector with the vector in the second position of the block. We then focus on the second vector, and search for its closest neighbor in the set of $(q-2)$ vectors. The process is then iterated. Details are given in \Cref{algo:permute}.

\subsection{Broader impacts and Safeguards}
\label{sec:supp_impacts}
Our computing unit seriously restricts the size of the decoder models we can train. We are not able to train one decoder model for a given LLM, but we work layer-wise and train a single decoder model for all patches extracted from the given layer. This layer-wise training forms the main limitation of \Algo\ \wrt\ standard post-training quantization methods, such as round to nearest (RTN).

This paper presents work whose goal is to advance the field of Machine Learning. There are many potential societal consequences of our work, none which we feel must be specifically highlighted here.

\begin{table}[ht]
  \caption{Perplexity $(\downarrow)$ on the validation dataset for LLaMA2-13B, with a sequence length of $2048$}
  \label{tab:perplexity_13b_ablation}
  \centering
  \begin{tabular}{cccc|c|c}
  \toprule
    Method & \#bits & rank $r$ & bucket $d$ & C4 $(\downarrow)$ & WikiText-2 $(\downarrow)$\\
    \midrule
      \Algo\ (no fine-tuning) & $3$ & $64$ & $2$ & $6.91$ & $5.27$ \\
      \Algo\ (30\% training) & $3$ & $64$ & $2$ & $6.69$ & $5.14$ \\
    \midrule
    \Algo\ (no fine-tuning) & $2$ & $64$ & $4$ & $10.36$ & $8.15$ \\
    \Algo\ (10\% training) & $2$ & $64$ & $4$ & $7.59$ & $5.99$ \\
    \bottomrule
  \end{tabular}
\end{table}

\end{document}